\documentclass[runningheads]{llncs}

\usepackage{eccv}

\usepackage{eccvabbrv}

\usepackage{amsmath}
\usepackage{multicol} % For multicolumn cells
\usepackage{colortbl}
\usepackage{longtable}
\usepackage{array}
\usepackage{arydshln}
\usepackage{multirow}
\usepackage{algorithm}
\usepackage[noend]{algpseudocode}
\usepackage{calligra}
\usepackage{mathrsfs}
\usepackage{tabulary}
\usepackage[normalem]{ulem} 

\usepackage{graphicx}
\usepackage{booktabs}

\usepackage[accsupp]{axessibility}  % Improves PDF readability for those with disabilities.

\usepackage{hyperref}

\usepackage{orcidlink}

\begin{document}

\title{ACTRESS: Active Retraining for Semi-supervised Visual Grounding}

\titlerunning{Active Retraining for Semi-supervised Visual Grounding}

\author{Weitai Kang\inst{1} \and
Mengxue Qu\inst{2} \and
Yunchao Wei\inst{2} \and
Yan Yan\inst{1}
}

\authorrunning{W.~Kang et al.}

\institute{Illinois Institute of Technology \and
Beijing Jiaotong University
}

\maketitle
\begin{abstract}
Semi-Supervised Visual Grounding (SSVG) is a new challenge for its sparse labeled data with the need for multimodel understanding.
A previous study, RefTeacher~\cite{sun2023refteacher}, makes the first attempt to tackle this task by adopting the teacher-student framework to provide pseudo confidence supervision and attention-based supervision.
However, this approach is incompatible with current state-of-the-art visual grounding models, which follow the Transformer-based pipeline. These pipelines directly regress results without region proposals or foreground binary classification, rendering them unsuitable for fitting in 
RefTeacher
due to the absence of confidence scores.
Furthermore, the geometric difference in teacher and student inputs, stemming from different data augmentations, induces natural misalignment in attention-based constraints. To establish a compatible SSVG framework, our paper proposes the {\bf ACT}ive {\bf RE}training approach for {\bf S}emi-{\bf S}upervised Visual Grounding, abbreviated as \textbf{ACTRESS}. Initially, the model is enhanced by incorporating an additional quantized detection head to expose its detection confidence. Building upon this, ACTRESS consists of an active sampling strategy and a selective retraining strategy. The active sampling strategy iteratively selects high-quality pseudo labels by evaluating three crucial aspects: \textit{Faithfulness}, \textit{Robustness}, and \textit{Confidence}, optimizing the utilization of unlabeled data. The selective retraining strategy retrains the model with periodic re-initialization of specific parameters, facilitating the model's escape from local minima. Extensive experiments demonstrates our superior performance on widely-used benchmark datasets.

\begin{figure}[t]
\centering
\includegraphics[width=0.82\textwidth]{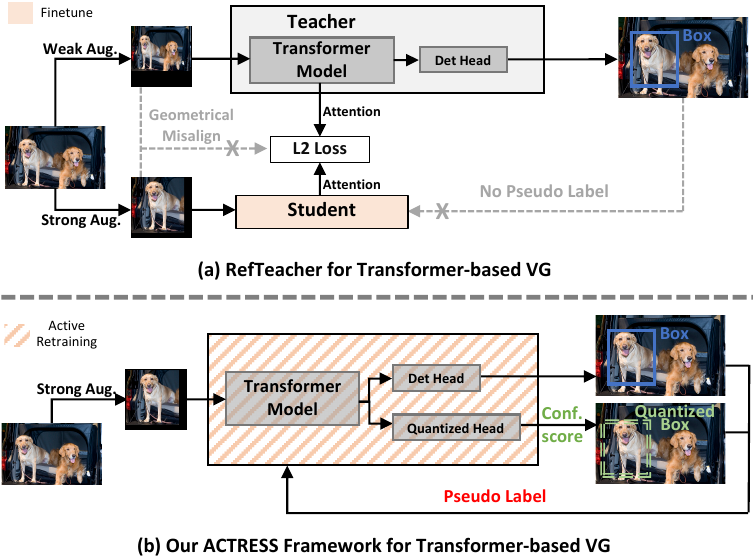}
\vspace{-6pt}
\caption{
Comparison the framework of our ACTRESS and RefTeacher when applied on SOTA (Transformer-based) models. We omit text inputs and labeled data.
\textbf{(a)} 
RefTeacher is incompatible since the absence of confidence score to provide pseudo label and geometrical misalignment in its attention loss.
\textbf{(b)}  
Instead, we equip the model with a quantized detection head to expose detection confidence, rendering them compatible with pseudo-label sampling. The Active Retraining helps escape local minima.
}
\label{intro}
\vspace{-10pt}
\end{figure}

\end{abstract}   
\section{Introduction}
Visual Grounding~\cite{kang2024segvg,kazemzadeh2014referitgame,mao2016generation,plummer2015flickr30k,yu2016modeling,kang2024intent3d} aims to localize a specific object within an image based on a free-form natural language expression. It is particularly important for numerous downstream multimodal reasoning systems, such as visual question answering~\cite{gan2017vqs,wang2020general,zhu2016visual7w,shang2024llava} and image captioning~\cite{anderson2018bottom,chen2020say,you2016image}. Previous models can be broadly categorized into three distinct groups: two-stage methods~\cite{yang2019dynamic,yang2020graph,yu2018mattnet,chen2021ref}, one-stage methods~\cite{yang2019fast,yang2020improving}, and Transformer-based ones~\cite{deng2021transvg,ye2022shifting,yang2022improving,qu2022siri,kang2024segvg,kang2024visual}.
Both two-stage and one-stage approaches rely on convolutional neural networks for region proposal or foreground binary classification, and manually-crafted techniques for language inference or multi-modal fusion.
Instead, current state-of-the-art (SOTA) methods follow the Transformer-based pipeline, which adopts the transformer structure to perform feature fusion to avoid manual bias, and directly regress the bounding box without candidate selection. 

Semi-supervised learning aims to address the costly labeling problem, which has been investigated in many computer vision tasks~\cite{bachman2014learning, berthelot2019remixmatch, berthelot2019mixmatch, gao2020consistency, liu2021unbiased, sohn2020fixmatch, sohn2020simple, tang2021humble, zhou2021instant}. 
A recent SSVG work, RefTeacher \cite{sun2023refteacher}, first notices this problem in Visual Grounding (VG). They adopt the teacher-student framework from semi-supervised object detection (SSOD), where the teacher model takes weakly augmented unlabeled data to stably provide pseudo confidence scores and attention features as supervision signals and the student is trained by strongly augmented data. 
However, this framework cannot be directly applied to SOTA VG models. As shown in Fig.\ref{intro}(a), these SOTA models adopt the Transformer-based pipeline to regress the result without region proposal or foreground binary classification, lacking the confidence score for pseudo label learning. 
Consequently, the only available constraint for unlabeled data is the L2 loss between the attention maps of the teacher and student models.
Moreover, an inherent issue arises because the teacher and student models process inputs with different data augmentations(RandomResize, RandomSizeCrop, NormalizeAndPad), leading to geometric misalignment of pixels in the input images. This misalignment makes the attention map constraint between the models less reliable. 
As a result, a superior Transformer-based model, TransVG \cite{deng2021transvg}, exhibits inferior performance in the semi-supervised setting under the RefTeacher framework \cite{sun2023refteacher}, compared with the less powerful one-stage model, RealGIN \cite{zhou2021real}.

To better harness the valuable information from unlabeled data and construct a semi-supervised framework compatible with Transformer-based VG models, we introduce in this paper the \textbf{ACT}ive \textbf{RE}training approach for \textbf{S}emi-\textbf{S}upervised visual grounding, \emph{a.k.a}, ACTRESS.

As shown in Fig.\ref{intro}(b), we simply sample new pseudo labels iteratively by our active sampling strategy
and retrain the model by the selective retraining strategy where certain parts of the model are re-initialized.
In this way, 
we can effectively make the best of the unlabeled data, and escape the local minima caused by training on limited data~\cite{qu2022siri}. 
Specifically, due to the lack of a pseudo label sampling mechanism,
previous works only use pseudo confidence scores for the detection loss and rely on limited labeled data for box regression learning to avoid the noise supervision from pseudo bounding boxes.
Instead, we aim to fully exploit the unlabeled data for box regression learning.
We first propose to equip the model with an additional quantized detection head, which is a simple MLP module. We then quantize the ground truth bounding box into discrete coordinates, which enables the head to learn detection ability by learning to classify the quantized coordinate. The model is then evaluated periodically to come up with pseudo labels. 
Based on the quantized detection head and the original detection head, 
our active sampling strategy selects pseudo labels from three aspects: \textit{Faithfulness}, \textit{Robustness}, and \textit{Confidence}. 
We supervise the box regression learning by both labeled data and our sampled high-quality pseudo labels. Since we only train one model with detection supervision, we avoid the above misalignment issue. 
Furthermore, we adopt the selective retraining strategy to escape the local minima of limited training samples, which re-initializes the backbones and the detection heads in each retrain stage. 
Unlike existing retraining approaches \cite{han2016dsd, qu2022siri} that ignore the quality and diversity of training samples, 
we perform our active sampling strategy to
re-sample new high-quality pseudo labels
at the beginning of each retrain stage and drop the old ones to avoid overfitting on the limited training set.

To validate our ACTRESS, we apply the method to one baseline Transformer-based model, TransVG \cite{deng2021transvg}, and the state-of-the-art one, VLTVG \cite{yang2022improving}.
Following the previous setting \cite{sun2023refteacher}, we conduct four protocols of semi-supervised learning experiments on three VG benchmark datasets, from 0.1\% labeled data to 10\% labeled data. Experimental results show that our ACTRESS can greatly exceed the supervised baselines and current SSVG framework, e.g. +11.84\% gains on 0.1\% RefCOCO. More importantly, using only 10\% labeled data, ACTRESS can help TransVG achieve 91.85\% fully supervised performance.

Our contributions are as follows: 
\textbf{(i)} We propose a new paradigm ACTRESS for semi-supervised visual grounding, which is more compatible with current SOTA VG models. 
\textbf{(ii)} We introduce a quantized detection head along with three reasonable pseudo label sampling strategies to make the best usage of unlabeled data.
\textbf{(iii)} We combine our active sampling strategy with the selective retraining strategy periodically to effectively escape from the local minima. 
\textbf{(iv)} We will release source code and checkpoints upon acceptance of the paper for future research development.

%% -------------------------------------------------------- %%
\section{Related Work}
\subsection{Visual Grounding}
Visual grounding approaches can be broadly categorized into three pipelines: two-stage methods, one-stage methods, and Transformer-based methods. Two-stage and one-stage methods primarily rely on CNN-based models.

\paragraph{\bf Two-stage / one-stage methods} 
Two-stage approaches~\cite{yu2018mattnet, chen2021ref} first generate candidate proposals and then select the one that has the best manually crafted matching score with the text.
One-stage methods~\cite{yang2019fast, yang2020improving} directly rank the confidence score of candidates which are proposed based on manually-designed multimodal features. 
However, both of the pipelines suffer from the manual bias to perform the query reasoning and multi-modal fusion~\cite{deng2021transvg}.

\paragraph{\bf Transformer-based methods} Current SOTA VG methods adopt the pipeline of Transformer-based approach, which is first introduced by TransVG~\cite{deng2021transvg}. They utilize Transformer~\cite{vaswani2017attention} encoders to perform cross-modal fusion among a learnable object query, visual tokens, and language tokens. The object query is then processed through an MLP module to predict the bounding box.
Benefiting from the flexible structure of Transformer modules in processing word-pixel level fusion without manual bias, recent works continue to adopt this pipeline and further propose novelties regarding feature extraction.
VLTVG~\cite{yang2022improving} comes up with a verification module before the decoder stage to encode the relationship between vision and language.
QRNet~\cite{ye2022shifting} proposes an early fusion of vision and text features to reduce the feature domain gap.

\subsection{Semi-Supervised Learning}
Semi-supervised Object Detection (SSOD) approaches primarily draw inspiration from semi-supervised image classification. 
A notable exemplar of SSOD is STAC \cite{sohn2020simple}, which initiates by training a teacher network using a limited set of annotated data and then generate pseudo-labels for unlabeled data. 
Following that, a student network is trained by both labeled and unlabeled data. 
Building upon this, several task-specific improvements have been proposed \cite{jeong2021interpolation, liu2021unbiased, liu2022unbiased, mi2022active, zhou2021instant} to consider the characteristics of object detection, where the teacher's stability is improved by EMA \cite{tarvainen2017mean} and weak-strong augmentation.
RefTeacher \cite{sun2023refteacher} then extends this SSOD paradigm to Semi-supervised Visual Grounding (SSVG), while they incorporate attention-based loss and confidence-based adaptive training for the sparse and noise pseudo label problem in visual grounding.

\subsection{Active Learning}
Several active learning methods have been proposed for label sampling in object detection \cite{wang2021data, yu2022consistency, yuan2021multiple, mi2022active}. For instance, CALD \cite{yu2022consistency} evaluates information based on data consistency. MI-AOD \cite{yuan2021multiple} utilizes multi-instance learning to mitigate pseudo-label noises. Active-Teacher \cite{mi2022active} extends the data sampling paradigm with a teacher-student framework. In this paper, our emphasis is on semi-supervised learning for visual grounding. Departing from the previous teacher-student learning framework, we instead adopt a selective retraining strategy.
\section{Proposed Method}

\begin{figure}[t]
    \centering
    \includegraphics[width=0.96\textwidth]{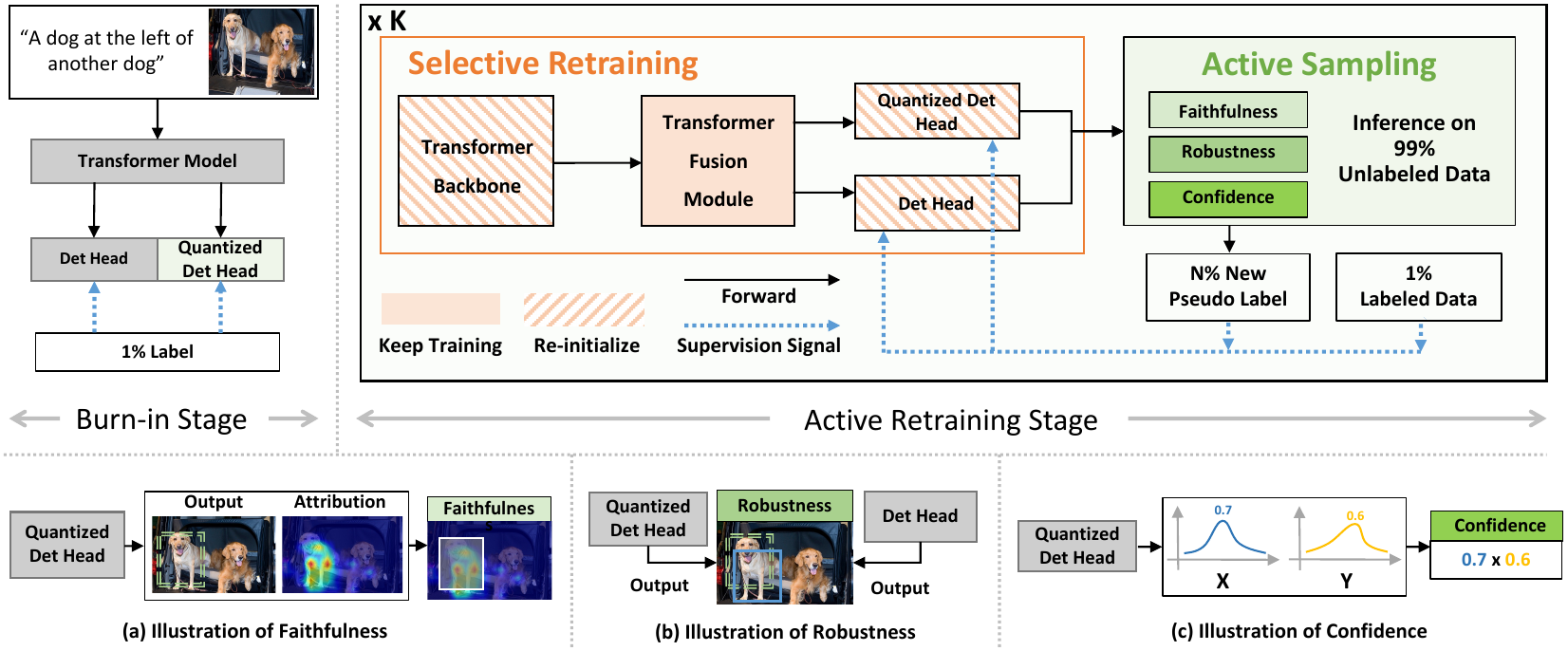}
    \caption{
    The overall framework of ACTRESS. Take 1\% labeled data as an example.
    {\bf Upper Left}: 
    We first train the model with 1\% labeled data, which is called the Burn-in stage. The model is equipped with an additional quantized detection head, which is supervised by a quantized ground truth bounding box.
    {\bf Upper Right}:
    In the Active Retraining Stage, the model performs inference on 99\% unlabeled data and evaluates three metrics: \textit{Faithfulness}, \textit{Robustness}, and \textit{Confidence}.
    Following that, we sample N\% new pseudo label with a high score of these three metrics.
    The sampled data together with 1\% labeled data go through geometrical transformation w/ or w/o quantization to supervise the two detection heads.
    Then the model selectively retrains the backbone and the detection heads, whose parameters are re-initialized at the beginning of the stage, and keeps training the fusion module. We periodically perform this stage for K times.
    {\bf Bottom}: Illustration of our three metrics.
    }
    \label{method}
\end{figure}

\begin{figure}[t]
    \centering % 
    \resizebox{0.96\textwidth}{!}{% 
    \begin{minipage}{\linewidth} % 
    \begin{algorithm}[H]
\caption{ACTRESS}\label{alg}
\textbf{Input:} 
Labeled data \(\{ ({\mathcal I^l}, {\mathcal T^l}), {\mathcal Y^l} \}\), 
unlabeled data \(\{ ({\mathcal I^u}, {\mathcal T^u}) \}\), 
Burn-in stage epoch $N_1$, Active Retraining stage epoch $N_2$,
iteration of Active Retraining stage $K$.
\begin{algorithmic}[1]
\For{i $< N_1$}

\item ~~~~~~Train the model on \(\{ ({\mathcal I^l}, {\mathcal T^l}), {\mathcal Y^l} \}\) by \ref{BI}
\EndFor
\For{m $< K$}

\item ~~~~~~Evaluate on \(\{ ({\mathcal I^u}, {\mathcal T^u}) \}\) to predict ${\mathcal Y^u}$

\item ~~~~~~Calculate three metrics on \(\{ ({\mathcal I^u}, {\mathcal T^u}), {\mathcal Y^u} \}\) by \ref{SSR}

\item ~~~~~~Sample pseudo label \(\{ ({\mathcal I^s}, {\mathcal T^s}), {\mathcal Y^s} \}\) by \ref{SSR}

\item ~~~~~~Re-initialize certain parts of the model by \ref{SSR}
~~~~ \For{j $< N_2$}

\item ~~~~~~~~~~~~Train the model on \(\{ ({\mathcal I^l} \textit{+} {\mathcal I^s} , {\mathcal T^l} \textit{+} {\mathcal T^s} ), {\mathcal Y^l} \textit{+} {\mathcal Y^s} \}\)
\EndFor

\EndFor
\end{algorithmic}
\end{algorithm}
\end{minipage}
}
\vspace{-10pt}
\end{figure}

\subsection{Overall Framework}

We begin by introducing the overarching framework of ACTRESS, designed for semi-supervised visual grounding. 
As shown in Fig.\ref{method}, 
there are two stages in our semi-supervised framework. The Burn-in stage is designed to initially train the model using labeled data, laying the foundation for subsequent pseudo-label generation. 
The main contribution of our ACTRESS is the Active Retraining stage, which consists of the active sampling strategy and selective retraining strategy.
The active sampling strategy is designed to harness the potential of unlabeled data through periodic pseudo-label sampling from three aspects. 
The selective retraining strategy retrains the model with the re-initialization of specific parameters at each Active Retraining stage, aiming to navigate away from local minima.
In summary, the entire ACTRESS procedure is outlined in Algorithm~\ref{alg}.

\subsection{Burn-in Stage}\label{BI}

The Burn-in stage initially trains the model with the labeled data, where the two detection heads develop the capability for subsequent pseudo-label prediction.
We mainly describe the quantized detection in this section.

\noindent \textbf{Quantized detection} Inspired from Pix2Seq~\cite{chen2021pix2seq}, we define Visual Grounding as a region classification task, where the ground truth is discretized into four (x, y, w, h) discrete coordinates. As shown in Fig.\ref{method}~(upper left), the quantized detection head, a simple MLP, is supervised to categorize which regions (discrete coordinates) the four values are situated into. 
For a more detailed understanding of the discretization process, we kindly encourage readers to refer to Pix2Seq~\cite{chen2021pix2seq}.

\subsection{Active Retraining Stage}\label{SSR}

As shown in Fig.\ref{method}~(upper right), we periodically perform the Active Retraining stage for K times, 
where our active retraining strategy first evaluates the model on the unlabeled data to propose pseudo labels, and then samples high-quality pseudo labels by assessing three aspects: \textit{Faithfulness}, \textit{Robustness}, and \textit{Confidence}. The training set of each stage consists of the new sampled pseudo label and the labeled data. Finally, we selectively retrain the model by re-initializing certain parts of the model.
We detail our three sampling aspects and the retraining strategy in the following sections.

\paragraph{\textbf{Faithfulness}} As argued by \cite{geirhos2020shortcut, chefer2022optimizing}, the model may remember certain shortcuts of the training set to some extent, especially the co-occurrence of the foreground (the target) and the background (the clue or shortcut), which can affect the model's generalization. In other words, the quality of the model's prediction deteriorates when the background changes in other datasets. 
Therefore, by assessing the extent to which the prediction is faithfully based on the foreground, we can evaluate its quality.
From this perspective, we define \textit{Faithfulness} as the degree of alignment between a prediction and its decision-making process to assess its fidelity to the foreground.
Here, to unveil the model's decision-making process, we employ TransMM~\cite{chefer2021generic}, which is an attribution heatmap calculating the contribution of each visual token towards forming the final prediction.
TransMM initials a diagonal matrix as its attribution map, and aggregates the attribution value of each token through the attention map by a forward pass using the update rules in Eq.\ref{relevance}: 
\vspace{-8pt}
\begin{equation}
\begin{aligned}
{\mathbf{R}^{vv}} = {\mathbf{R}^{vv}} + {\overline{\mathbf{A}}} \cdot {\mathbf{R}^{vv}} \\
{\mathbf{R}^{rv}} = {\mathbf{R}^{rv}} + {\overline{\mathbf{A}}} \cdot {\mathbf{R}^{vv}} 
\end{aligned}
\label{relevance}
\vspace{-8pt}
\end{equation}
where $\mathbf{R}^{vv}$ denotes the attribution values from visual tokens to visual tokens.
$\mathbf{R}^{rv}$ is the attention map weighted summed by the classification result's gradient of each head through a ReLU layer.
Due to space constraints, we omit certain details such as normalization and kindly refer the readers to \cite{chefer2021generic} for further information.
As we have equipped the model with an additional quantized detection head for location classification, 
we can compute the gradient based on the summation of the argmax classification results of the four values (x, y, w, h).
Finally, the attribution map is upsampled to image resolution. As shown in Fig.\ref{method}~(bottom), \textit{Faithfulness} is then calculated as the ratio of the summation of attribution values within the predicted box to the summation of the entire attribution map. Higher \textit{Faithfulness} scores indicate greater fidelity to the foreground, signifying higher-quality pseudo labels. Here, it should be noted that in some cases a slight focus on background may help in VG. Yet, the main problem in SSVG setting is filtering good results from abundant noisy outputs, which totally miss the foreground, overshadowing the nuance of considering the background. We will show the effectiveness of our \textit{Faithfulness} in the experiments.

\paragraph{\textbf{Robustness}} A more discriminative object query exhibits a higher likelihood of accurate decoding into the bounding box result and demonstrates better robustness when applied across various decoding perspective. Specifically, when comparing two distinct detection formats - the regression output from the original detection head and the classification output from the quantized detection head – we anticipate that the two detection outputs produced by their respective heads will coincide, as they decode the same object query. To evaluate this consistency, as shown in Fig.\ref{method}~(bottom), we employ the Generalized Intersection over Union (IoU) metric~\cite{rezatofighi2019generalized} to quantify the degree of concurrence between these two detection outputs, which serves as the measure of robustness. This approach enables us to generate higher-quality pseudo labels by sampling the pseudo labels with higher \textit{Robustness} scores.

\paragraph{\textbf{Confidence}} The confidence value in a model's prediction is often used as a metric to gauge the quality of pseudo labels~\cite{sun2023refteacher, mi2022active}, as it is learning to be as high as possible to signify model's confirmation in distinguishing the foreground. 
However, Transformer-based models in visual grounding inherently lack confidence values, as they directly regress bounding box coordinates without providing a measure of confidence. 
Instead, through the integration of an additional quantized detection head, we can successfully infer confidence values for the model's predictions.
Significantly, this confidence measure differs from conventional foreground classification confidence; instead, it pertains to the confidence of predicted values of the bounding box, specifically reflecting location certainty. This aligns more closely with our need to select high-quality bounding boxes for regression constraints.
As shown in Fig.\ref{method}~(bottom), using the quantized detection output, we combine the maximum confidence values of the x and y coordinates to compute the confidence score for our model's detection, considering that 
position (xy) is easier and steadier to predict than including
shape (xywh).

Finally, our active sampling strategy assesses a composite measure that takes into account all of these three aspects to select high-quality pseudo labels in each Active Retraining stage.
We define the sampling metric as follows:
\begin{equation}
\begin{aligned}
{\mathrm{I_{Act}}} = {\mathrm{Norm(I_f)}} \cdot {\mathrm{Norm(I_r)} \cdot {\mathrm{Norm(I_c)}}} \\
\end{aligned}
\label{combination}
\end{equation}
where $\mathrm{Norm}$ denotes the min-max normalization of all pseudo labels, $\mathrm{I_f}$ represents the Faithfulness score of each sample, $\mathrm{I_r}$ signifies the Robustness score of each sample, $\mathrm{I_c}$ stands for the Confidence score of each sample, and $\mathrm{I_{Act}}$ is the final metric, which combines these three factors of each sample. We sample the top N\% pseudo labels at the beginning of this stage.

\paragraph{\bf{Selective Retraining}} Transformer-based models often face overfitting challenges in semi-supervised learning due to the limited number of available labels. To mitigate this problem, we employ two key approaches: introducing new pseudo labels and implementing selective retraining.
In the initial stage of the Active Retraining process, we refresh the pseudo labels by re-sampling and discarding the previous ones. 
Subsequently, we re-initialize the backbone and detection heads of the model, similar with~\cite{qu2022siri}. This step aims to reduce the risk of overfitting by adjusting model parameters while preserving the fusion capability of the fusion module. 
Through these two ways, we alleviate the overfitting issue from the perspectives of both the training data and the model's parameters. 
\section{Experiments}

\subsection{Dataset and Evaluation}
\paragraph{\bf Dataset} 
Following previous work~\cite{sun2023refteacher}, 
we evaluate our method on three widely used datasets: RefCOCO~\cite{yu2016modeling}, RefCOCO+~\cite{yu2016modeling}, and RefCOCOg-umd~\cite{mao2016generation}, with four semi-supervised learning setting: 0.1\% labeled data, 1\% labeled data, 5\% labeled data, and 10\% labeled data. 
The rest are used as unlabeled data.

\begin{table*}[t]
\centering
\caption{\textbf{Quantitative comparisons with the previous methods on RefCOCO, RefCOCO+ and RefCOCOg-umd.} STAC and RefTeacher use RealGIN~\cite{zhou2021real} as the VG model, and their performance is referenced from~\cite{sun2023refteacher}. Our ACTRESS is applied to two Transformer-based models: TransVG~\cite{deng2021transvg} and VLTVG~\cite{yang2022improving}.
}
\vspace{-10pt}
\resizebox{1\linewidth}{!}{
\setlength{\tabcolsep}{0.6mm}
\begin{tabular}{ccccccccccccc}
\specialrule{1.4pt}{1pt}{1pt} % 
\multirow{2}{4em}{Methods} & \multicolumn{12}{c}{RefCOCO} \\
\cline{2-13}
& \multicolumn{3}{c}{0.1\%} & \multicolumn{3}{c}{1\%} & \multicolumn{3}{c}{5\%} & \multicolumn{3}{c}{10\%}\\
\cline{1-13}
&  val & testA & testB &  val & testA & testB &  val & testA & testB &  val & testA & testB \\
\cmidrule(lr){2-4} \cmidrule(lr){5-7} \cmidrule(lr){8-10} \cmidrule(lr){11-13}
STAC~\cite{sohn2020simple} &18.68 &22.11 &14.27 &43.69 &47.75 &37.29& 58.14 &59.64 &53.64 &62.35& 64.56& 58.06 \\
RefTeacher~\cite{sun2023refteacher}& 34.05& 35.41& 30.25& 59.25& 60.47& 56.11& 68.96& 71.04& 63.18 &72.22 &74.47& 66.69 \\
ACTRESS (TransVG) & \textcolor{blue}{38.37} & \textcolor{blue}{43.18} & \textcolor{blue}{34.99} & \textcolor{blue}{59.86} & \textcolor{blue}{62.56} & \textcolor{blue}{57.10} & \textcolor{blue}{71.62} & \textcolor{blue}{74.84} & \textcolor{red}{68.47} & \textcolor{blue}{76.26} & \textcolor{blue}{78.69} & \textcolor{red}{71.76} \\
ACTRESS (VLTVG) & \textcolor{red}{45.42} & \textcolor{red}{49.96} & \textcolor{red}{39.84} & \textcolor{red}{62.47} & \textcolor{red}{65.08}& \textcolor{red}{58.67} & \textcolor{red}{72.81}  & \textcolor{red}{75.08}  & \textcolor{blue}{68.25} & \textcolor{red}{76.90} & \textcolor{red}{79.23} & \textcolor{blue}{71.74} \\
\specialrule{0.8pt}{1pt}{1pt}
\specialrule{0.8pt}{1pt}{1pt}
\multirow{2}{4em}{Methods} & \multicolumn{12}{c}{RefCOCO+} \\
\cline{2-13}
& \multicolumn{3}{c}{0.1\%} & \multicolumn{3}{c}{1\%} & \multicolumn{3}{c}{5\%} & \multicolumn{3}{c}{10\%}\\
\cline{1-13}
&  val & testA & testB &  val & testA & testB &  val & testA & testB &  val & testA & testB \\
\cmidrule(lr){2-4} \cmidrule(lr){5-7} \cmidrule(lr){8-10} \cmidrule(lr){11-13}
STAC~\cite{sohn2020simple} & 17.96 & 22.13 & 11.45 & 30.48 & 32.68 & 23.71 & 38.88 & 40.97 & 31.70 & 40.83 & 43.42 & 34.51 \\
RefTeacher~\cite{sun2023refteacher}& 22.10 & 28.66 & 12.76 & \textcolor{blue}{39.45} & 41.95 & 32.17 & 49.47 & 53.27 & 41.52 & 52.50 & 56.76 & 44.69 \\
ACTRESS (TransVG) & \textcolor{blue}{29.96} & \textcolor{blue}{32.35} & \textcolor{blue}{26.47} & 38.42 & \textcolor{blue}{42.58} & \textcolor{blue}{32.95} & \textcolor{blue}{49.67} & \textcolor{blue}{55.88} & \textcolor{blue}{42.62} & \textcolor{blue}{56.05} & \textcolor{blue}{62.55} & \textcolor{blue}{47.57} \\
ACTRESS (VLTVG) & \textcolor{red}{31.24} & \textcolor{red}{33.92} & \textcolor{red}{27.34} & \textcolor{red}{40.71} & \textcolor{red}{43.97}& \textcolor{red}{35.97} & \textcolor{red}{51.90}  & \textcolor{red}{57.63}  & \textcolor{red}{43.03} & \textcolor{red}{58.01} & \textcolor{red}{64.57} & \textcolor{red}{49.59} \\
\specialrule{0.8pt}{1pt}{1pt}
\specialrule{0.8pt}{1pt}{1pt}
\multirow{2}{4em}{Methods} & \multicolumn{12}{c}{RefCOCOg-umd} \\
\cline{2-13}
& \multicolumn{3}{c}{0.1\%} & \multicolumn{3}{c}{1\%} & \multicolumn{3}{c}{5\%} & \multicolumn{3}{c}{10\%} \\
\cline{1-13}
&  val-u & \multicolumn{2}{c}{test-u} &  val-u & \multicolumn{2}{c}{test-u} &  val-u & \multicolumn{2}{c}{test-u}&  val-u & \multicolumn{2}{c}{test-u} \\
\cmidrule(lr){2-4} \cmidrule(lr){5-7} \cmidrule(lr){8-10} \cmidrule(lr){11-13}
STAC~\cite{sohn2020simple} &  18.28 &  \multicolumn{2}{c}{18.18 }&  31.52 & \multicolumn{2}{c}{ 30.77} &  43.18 &  \multicolumn{2}{c}{41.65} &  47.86 &  \multicolumn{2}{c}{47.74} \\
RefTeacher~\cite{sun2023refteacher}& 29.06 & \multicolumn{2}{c}{29.76} & \textcolor{blue}{44.02} & \multicolumn{2}{c}{42.13} & 51.20 & \multicolumn{2}{c}{50.79} & 51.20 & \multicolumn{2}{c}{56.80} \\
ACTRESS (TransVG) & \textcolor{blue}{32.13} & \multicolumn{2}{c}{\textcolor{blue}{31.27}} & 43.42 & \multicolumn{2}{c}{\textcolor{blue}{42.78}} & \textcolor{blue}{56.17}& \multicolumn{2}{c}{\textcolor{blue}{55.09}} & \textcolor{blue}{61.40} & \multicolumn{2}{c}{\textcolor{blue}{60.67}}\\
ACTRESS (VLTVG) & \textcolor{red}{33.03} & \multicolumn{2}{c}{\textcolor{red}{33.22}} & \textcolor{red}{49.82} & \multicolumn{2}{c}{\textcolor{red}{47.92}} & \textcolor{red}{58.88}& \multicolumn{2}{c}{\textcolor{red}{58.07}} & \textcolor{red}{61.89} & \multicolumn{2}{c}{\textcolor{red}{60.98}}\\
\specialrule{1.4pt}{1pt}{1pt} % 
\end{tabular}
}
\label{table1}
\end{table*}

Images of these datasets are selected from MSCOCO \cite{lin2014microsoft}. 
RefCOCO \cite{yu2016modeling} has 19,994 images with 50,000 referred objects and 142,210 referring expressions. It is officially split into four datasets: training set with 120,624 expressions, validation set with 10,834 expressions, testA set with 5,657 expressions, and testB set with 5,095 expressions.
RefCOCO+ \cite{yu2016modeling} is a harder benchmark since its language is not allowed to include location words but is just allowed to contain purely appearance-based descriptions. It has 19,992 images with 141,564 referring expressions for 49,856 referred objects. It is split into four datasets: training set with 120,191 expressions, validation set with 10,758 expressions, testA set with 5,726 expressions, and testB set with 4,889 expressions. RefCOCOg-umd~\cite{mao2016generation} is also a harder benchmark as it contains a large number of hard cases for its flowery and complex expressions. Especially, the length of its language is longer than that of others. It has 25,799 images with 49,822 object instances and 95,010 expressions and is split into a training set, a validation set and a test set.

\noindent {\bf Evaluation} Following the previous setting~\cite{sun2023refteacher,deng2021transvg,yang2022improving}, we use the top-1 accuracy(\%) to evaluate our method, where the predicted bounding box is regarded as positive if its IoU with the ground-truth bounding box is greater than 0.5. 

\subsection{Implementation Details}
We verify the effectiveness of ACTRESS by applying it to two representative Transformer-based models: the baseline one TransVG~\cite{deng2021transvg} and the state-of-the-art one VLTVG~\cite{yang2022improving}. We follow their original training configuration, including optimizer, learning rate and so on. We specify the batchsize, epoch, and iteration in Supplementary.
The total iteration is roughly comparable to the configuration in RefTeacher. 
It is noted that the ratio of labeled data to unlabeled data is maintained at 3:1 in each batch, by which we ensure the labeled data dominates the batch for a high quality training batch.
For the x\% labeled data setting, we sample x\% of the unlabeled data in each Active Retraining stage.

\subsection{Quantitative Results}

In Tab.\ref{table1}, we compare ACTRESS with STAC\cite{sohn2020simple} and RefTeacher~\cite{sun2023refteacher}. ACTRESS is applied to two Transformer-based models, while previous methods use a one-stage model~\cite{zhou2021real}.
The first observation is that all our ACTRESS models achieve a new state-of-the-art performance in semi-supervised visual grounding (SSVG). Due to the compatible design of ACTRESS, we successfully extend the achievements of current VG models to the semi-supervised learning scenario.
Compared with RefTeacher, both the baseline Transformer-based model, TransVG, and the state-of-the-art one, VLTVG, achieve superior performance under our ACTRESS framework on most of the SSVG benchmarks.
On average, ACTRESS (TransVG) achieves a +7.04\% improvement on 10\% RefCOCOg-umd and +8.42\% on 0.1\% RefCOCO+. ACTRESS (VLTVG) achieves a +9.66\% improvement on 0.1\% RefCOCO+ and +11.84\% on 0.1\% RefCOCO.
Specifically, the performance gains by ACTRESS (VLTVG) are the most significant, e.g., +11.37\% and +14.55\% on 0.1\% RefCOCO $val$ and $testA$; +14.58\% on 0.1\% RefCOCO+ $testB$; +10.69\% on 10\% RefCOCOg-umd $val$-$u$. 
The second observation is that the comparison between ACTRESS (TransVG) and ACTRESS (VLTVG) shows that ACTRESS (VLTVG) almost uniformly outperforms ACTRESS (TransVG). This indicates that ACTRESS effectively leverages the advancements of the model itself, demonstrating that a superior model yields better performance in semi-supervised learning within our framework, which is not the case in RefTeacher~\cite{sun2023refteacher}, as discussed in the Introduction section.

\subsection{Ablation Study}

In this section, we conduct a series of detailed ablation studies, which include controlling the model used to compare different SSVG frameworks and evaluating the effectiveness of each of our proposed ideas.

\paragraph{\bf Controlling the model as TransVG} As shown in Tab.~\ref{ablation1}, we control the variable of the model we used, i.e., TransVG, to compare ACTRESS with RefTeacher on the validation sets. We observed an impressive improvement, i.e., +6.0\% on RefCOCO, +9.7\% on RefCOCO+, and +10.4\% on RefCOCOg-umd. Such improvements justify our compatibility with current advanced VG models.
Moreover, when we compute the average ratio of our performance on each dataset relative to the performance achieved through fully supervised learning, our ACTRESS helps TransVG achieve 91.85\% of the full training performance, using only 10\% labeled data.

\begin{table}[t]
    \centering
    \caption{\textbf{Quantitative comparisons while controlling the model (TransVG).} RefTeacher and ACTRESS are trained on 10\% labeled data. ``Full'' indicates the performance of fully supervised learning using 100\% labeled data.}
    \vspace{-10pt}
    \setlength{\tabcolsep}{3mm}
    \begin{tabular}{cccc}
        \toprule
        Methods & RefCOCO & RefCOCO+ & RefCOCOg-umd \\
        \midrule
        RefTeacher (TransVG) & 70.3 & 46.4 & 51.0 \\
        \rowcolor{gray!30} ACTRESS (TransVG) & 76.3 & 56.1 & 61.4 \\
        \rowcolor{gray!50} Full (TransVG) & 80.3 & 63.5 & 66.6 \\
        \bottomrule
      \end{tabular}
    \label{ablation1}
    \vspace{-8pt}
\end{table}

\begin{table}[t]
\centering
\caption{\textbf{Ablation study on VLTVG.} The model is trained on 10\% labeled data.
}
\vspace{-10pt}
\setlength{\tabcolsep}{2.4mm}
\begin{tabular}{cccc}
\specialrule{.8pt}{1pt}{1pt}
Methods & RefCOCO & RefCOCO+ & RefCOCOg-umd \\
\specialrule{.4pt}{1pt}{1pt}
RefTeacher (RealGIN)~\cite{sun2023refteacher} & 72.2 & 52.5 & 56.5 \\
\specialrule{.4pt}{1pt}{1pt}
\rowcolor{gray!10} Supervised (VLTVG) & 73.99 & 54.22 & 59.91 \\
\specialrule{.4pt}{1pt}{1pt} % 
\rowcolor{gray!30} ACTRESS (VLTVG) & 76.90 & 58.01 & 61.89 \\
\specialrule{.8pt}{1pt}{1pt}
\end{tabular}
\label{ablation2}
\end{table}

\begin{table}[t]
\centering
\caption{\textbf{Quantitative comparisons with Siri~\cite{qu2022siri}.} We use TransVG as the model, which is trained on 1\% and 10\% labeled data, respectively.
}
\setlength{\tabcolsep}{3mm}
\begin{tabular}{ccc}
\specialrule{.8pt}{1pt}{1pt}
Methods & 1\% RefCOCOg-umd & 10\% RefCOCOg-umd \\
\specialrule{.4pt}{1pt}{1pt}
RefTeacher~\cite{sun2023refteacher} & 35.17	& 51.0 \\
\specialrule{.4pt}{1pt}{1pt} % 
\rowcolor{gray!10} Siri~\cite{qu2022siri} & 39.48	& 57.48   \\
\specialrule{.4pt}{1pt}{1pt} % 
\rowcolor{gray!30} ACTRESS & 43.42 & 61.40  \\
\specialrule{.8pt}{1pt}{1pt}
\end{tabular}
\label{ablation3}
\end{table}

\begin{table}[t]
    \centering
    \caption{\textbf{Evaluate the effectiveness of each sampling metric.} We train TransVG by the Burn-in stage on 10\% RefCOCOg-umd.
    }
    \setlength{\tabcolsep}{2mm}
    \begin{tabular}{cccccc}
    \specialrule{.8pt}{1pt}{1pt}
    Metrics & Top-50\% & Top-40\% & Top-30\% & Top-20\% & Top-10\%   \\
    \specialrule{.4pt}{1pt}{1pt}
    Random & 49.57 & 49.44 & 49.01 & 50.05 & 49.83 \\
    \specialrule{.4pt}{1pt}{1pt} %
    Robust(R) & 57.05 & 57.97 & 58.89 & 60.07 & 61.5 \\
    \specialrule{.4pt}{1pt}{1pt} % 
    Conf(C) & 59.06 & 60.39 & 61.45 & 62.57 & 64.35 \\
    \specialrule{.4pt}{1pt}{1pt}
    Faith(F) & 57.12 & 59.25 & 62.2 & 65.86 & 70.48 \\
    \specialrule{.4pt}{1pt}{1pt}
    \rowcolor{gray!30} R*C*F & 60.58 & 62.54 & 64.93 & 67.72 & 70.59 \\
    \specialrule{.8pt}{1pt}{1pt}
    \end{tabular}
    \label{ablation4}
    \end{table}
    
\begin{table}[t]
    \centering
    \caption{\textbf{Evaluate the improvement gained from pseudo labels.} We train TransVG by the Burn-in stage on 10\% RefCOCOg-umd and one round of the Active Retraining stage on the sampled data and labeled data.
    }
    \setlength{\tabcolsep}{3mm}
    \begin{tabular}{ccccc}
    \specialrule{.8pt}{1pt}{1pt}
    \multirow{2}{*}{Methods} &  \multicolumn{3}{c}{Metrics}  & \multirow{2}{*}{RefCOCOg-umd}   \\
    \cline{2-4}
     & Robust & Conf & Faith &  \\
    \specialrule{.4pt}{1pt}{1pt}
    RefTeacher~\cite{sun2023refteacher} & - & - & - & 51.00 \\
    \specialrule{.4pt}{1pt}{1pt} % 
    Robust(R) & \checkmark & - & - & 56.00 \\
    Conf(C) & - & \checkmark & - & 56.96 \\
    Faith(F) & - & - & \checkmark & 56.50 \\
    \rowcolor{gray!30} R*C*F & \checkmark & \checkmark & \checkmark & 57.25 \\
    \specialrule{.8pt}{1pt}{1pt}
    \end{tabular}
    \label{ablation5}
\end{table}

\begin{table}[t]
    \centering
    \caption{Results on ReferItGame and Flickr30k with TransVG and 10\%labeled data. ACTRESS$^*$ means we stop training once we exceeds the performance of RefTeacher.
    }
    \setlength{\tabcolsep}{3mm}
    \begin{tabular}{lcccccc}
    \hline
    \multirow{2}{*}{Methods} & \multicolumn{3}{c}{ReferItGame} & \multicolumn{3}{c}{Flickr30k} \\ 
    \cline{2-7}
     & val & test & time cost & val & test & time cost \\ 
    \hline
    RefTeacher & 48.7 & 46.9 & 10.3h & 61.5 & 61.7 & 9.5h \\
    \rowcolor{gray!10} ACTRESS$^*$ & 49.3 & - & 3.0h & 62.1 & - & 1.1h \\
    \rowcolor{gray!30} ACTRESS & 56.4 & 54.2 & 10.0h & 67.9 & 68.4 & 9.2h \\
    \hline
    \end{tabular}
    \label{tab7}
\end{table}

\paragraph{\bf Controlling the model as VLTVG} To further verify the effectiveness of ACTRESS and the motivation of adopting SOTA VG model, we compare our ACTRESS (VLTVG) with the supervised learning method, for which we follow the default setup of VLTVG to train the model on available labeled data.
As shown in the second and third lines of Tab.~\ref{ablation2}, we consistently achieve improvements on all three benchmarks. When comparing the results of the first and second line of Tab.~\ref{ablation2}, it is worth noting that even with simple supervised learning on VLTVG, the performance has already surpassed the previous SSVG method, RefTeacher, which is only compatible with one-stage VG models. Therefore, it is necessary for us to develop a compatible SSVG framework to fully leverage the advancements of current Transformer-based models.

\paragraph{\bf Impact of active sampling strategy} To demonstrate the effectiveness of our proposed active sampling strategy, we conducted an ablation study where we implemented selective retraining without introducing new pseudo labels. This approach is akin to Siri~\cite{qu2022siri} in a full-training setting, but in our case, it is trained on the limited labeled data. As shown in Tab.\ref{ablation3}, simply retraining specific parts of the model can effectively make the best of the labeled data and surpass RefTeacher, which utilizes all of the unlabeled data for training. However, this retraining strategy still faces the challenge of overfitting due to the extremely limited labeled data in SSVG. In contrast, when we incorporate our active sampling strategy, the training set is periodically extended and refreshed, effectively addressing the overfitting issue and resulting in a significant improvement across all the benchmarks in Tab.\ref{ablation3}.

\paragraph{\bf Impact of each sampling metric} Concerning the sampling metrics, we assess the effectiveness of each individual metric and their final combination from two aspects: 
the quality of the sampled pseudo labels and the improvement achieved by training on the labeled data in conjunction with the sampled data.

\begin{figure*}[t]
\centering
\includegraphics[width=1\textwidth]{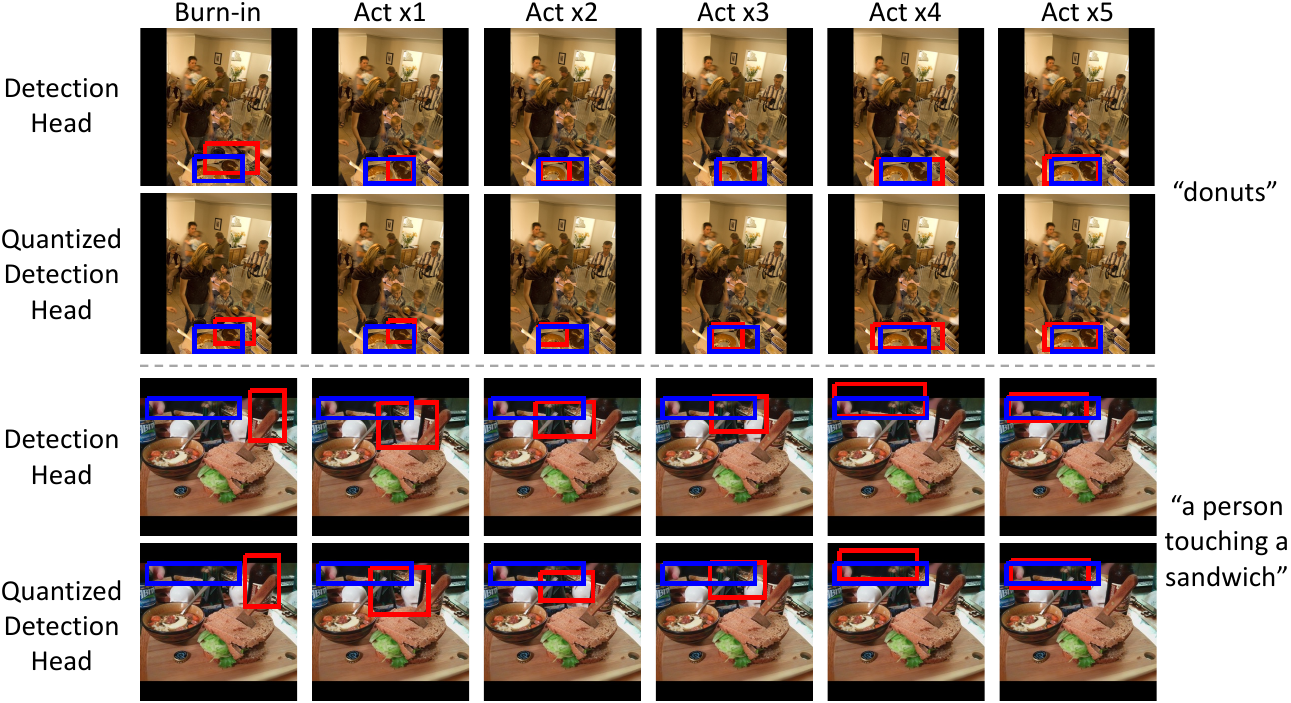}
\caption{Predictions from two detection heads. The blue box is the ground truth and the red box is the prediction. We compare the prediction quality of the model trained after the Burn-in stage and after multiple rounds of the Active Retraining stage (Act). It is evident that the quality improves consistently with an increase in the number of rounds of the Active Retraining stage. Texts on the right side are the input text.}
\label{draw}
\end{figure*}

Firstly, we train TransVG with the Burn-in stage on 10\% RefCOCOg-umd. Subsequently, we conduct an evaluation on the remaining 90\% unlabeled data to generate pseudo labels. We assess the effectiveness of each metric by ranking the quality of the pseudo labels according to each individual metric, as well as the final combination. We evaluate the quality of the sampled data by testing the accuracy of the top-k\% pseudo labels.
As shown in Tab.\ref{ablation4}, when using a random index to rank the pseudo labels, the quality of the sampled data does not improve as we increase the threshold from top-50\% to top-10\%. In contrast, each of the sampling metrics—\textit{Faithfulness}, \textit{Robustness}, and \textit{Confidence}—can effectively serve as a sampling index, as the accuracy increases when we raise the threshold. Ultimately, the combination of these three metrics using Eq.~\ref{combination} produces the best sampling index for high-quality pseudo labels, which verifies the effectiveness of Eq.~\ref{combination} in considering all metrics together.

Secondly, we train the model on both the labeled data and the sampled data for one round of the Active Retraining stage. As indicated in Tab.~\ref{ablation5}, when compared with RefTeacher, each set of sampled data effectively improves performance to surpass RefTeacher. Ultimately, the combination one achieves the best results, which verifies the necessity of considering all three metrics spontaneously.

\subsection{Performance on other image sources}
Previous study evaluates performance only on benchmarks with images from MSCOCO~\cite{lin2014microsoft}. To further complement the SSVG benchmark with more different data sources to align with VG benchmarks, we conduct additional experiments on ReferItGame dataset~\cite{kazemzadeh2014referitgame}, whose images are from SAIAPR-12~\cite{escalante2010segmented}, and on Flickr30k Entities dataset (also abbreviated as Flickr30k)~\cite{plummer2015flickr30k}, whose images are from Flickr30k~\cite{young2014image}. All experiments are conducted within 75k iterations. As shown in the first and third line of Tab.~\ref{tab7}, our ACTRESS excels RefTeacher by +7.7\%, +7.3\% on ReferItGame and +6.4\%, +6.7\% on Flickr30k, respectively. 

\subsection{Comparison regarding time cost}
Since ACTRESS only trains one model on selected data with periodically (K times, K $<<$ training epoch) evaluating the three metrics, our method possesses the virtue of efficiency. In contrast, RefTeacher updates two models (teacher and student) on all the unlabeled data and labeled data with two sets of data augmentation.
To demonstate the efficiency of our ACTRESS, we compare the training time cost of our ACTRESS with RefTeacher. 
As shown in Tab.~\ref{tab7}, we achieve a comparable performance with RefTeacher using only 29.1\% and 11.6\% of the time cost of RefTeacher on ReferItGame and Flickr30k, respectively. When using the same time cost, our ACTRESS excels RefTeacher by a large margin.

\subsection{Qualitative Results}
In this section, we visualize the prediction results generated by the two detection heads. 
We train TransVG on 10\% RefCOCOg-umd and compare the  predictions in the Burn-in stage and several rounds of the Active Retraining stage.
As depicted in the upper two lines of Fig.~\ref{draw}, a simple case with ``donuts'' text input, the model can roughly locate the target. After multiple iterations of Active Retraining, the model gradually escapes from local minima by incorporating newly sampled data and a retraining strategy. Consequently, both of the two detection heads refine the prediction to correctly ground the ``donuts''. 
Another case in the lower two lines of Fig.~\ref{draw} indicates a harder case with longer input text. The model completely misses the target after the initial training of the Burin-in stage. 
This inaccurate case could introduce noise signals if utilized as a pseudo label.
However, our ACTRESS can exclude those low-quality pseudo labels
and help the model progressively improve the detection ability.
The improved pseudo labels can further enhance the model's learning in subsequent rounds.
Eventually, both of the two detection heads achieve accurate prediction results.

\section{Conclusion}
In conclusion, we provide ACTRESS for semi-supervised visual grounding, compatible with SOTA VG models. Specifically, we propose a quantized detection head and an active sampling strategy to maximize the utilization of unlabeled data. Furthermore, our approach combines the sampling strategy with a selective retraining mechanism, aiding in the escape from local minima during training. Through extensive experiments across four semi-supervised learning protocols on three VG benchmark datasets, we empirically demonstrate the superior performance of our proposed methods compared to previous SOTA techniques.

% \clearpage  % TODO REVIEW/FINAL: This \clearpage needs to be removed from both review and camera-ready versions.

\bibliographystyle{splncs04}
\bibliography{main}
\clearpage
\begin{center}
    \centering
    \captionsetup{type=figure}
    \includegraphics[width=.93\textwidth]{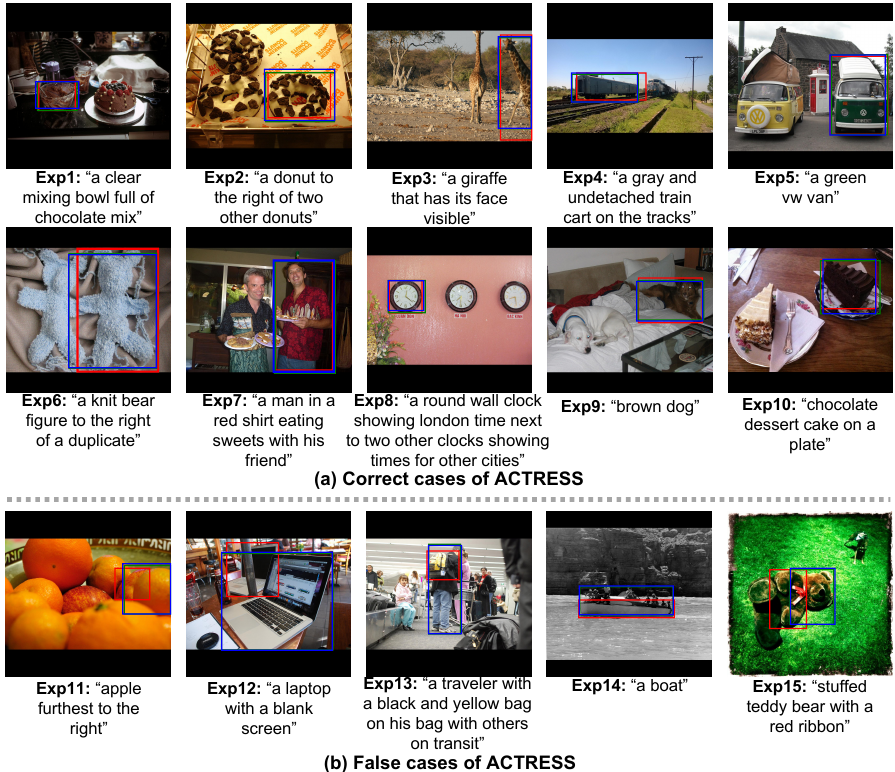}
    \captionof{figure}{Visualization of correct cases and false cases of ACTRESS. Red boxes are ground truth boxes, green boxes are detection outputs, and blue boxes are quantized detection outputs.}
    \label{sup_vis}
\end{center}%

\section{Additional Qualitative Results}
As shown in Fig.~\ref{sup_vis}, we provide more qualitative results of our ACTRESS, including correct and false cases. In the upper two rows (Fig.~\ref{sup_vis} (a)), spanning from Exp1 to Exp10, ACTRESS shows robustness across all cases where there are distracting candidate objects in the image. In the bottom row (Fig.~\ref{sup_vis} (b)), we also present cases where our ACTRESS fails. Specifically, as shown in Exp11 and Exp12, our semi-supervised learning still struggles to completely address certain occlusion cases. Additionally, we identified ambiguity issues in the annotations, leading to failed cases. For example, in Exp14, the ground truth box (in red) and predictions (in blue and green) from ACTRESS can all be considered correct. The ground truth box focuses on the main body of the "boat", while ours includes the entire body of the "boat". Furthermore, there are some mistakes in annotation, illustrated in Exp13, where the final target of the text is on the "traveler", and our prediction is accurate, but the annotation mistakenly references the "bag". Exp15 serves as another example of annotation error, where the annotation should crop the "bear" instead of the "ribbon", causing ambiguity in our learning process.

\begin{table}[t]\scriptsize
\centering
\caption{Setting of RefCOCO. 
"PR" indicates the percentage of labeled data. 
"BS" is the training size of one batch.
"BI~Ep" is the number of epoch of the Burin-in stage.
"ACT~Ep" is the number of epoch of each Active Retraining stage.
"ACT~It" is the iteration of ACT stage.
"LRD~Ep" refers to the epoch at which the learning rate starts to decrease to one-tenth of its value.
"Tr~St" is the total number of training step.
}
% \vspace{-10pt}
\setlength{\tabcolsep}{3mm}
\begin{tabular}{c|c|c|c|c|c|c}
\hline
PR & BS & BI~Ep & ACT~Ep & ACT~It & LRD~Ep & Tr~St  \\
\hline
$0.1\%$ & 8 &  400 & 150 & 11 & 2000 & 105.00k \\
\hline
$1\%$ & 16 & 80 & 30 & 11 & 400 & 104.67k\\
\hline
$5\%$ & 32 & 30 & 30 & 5 & 170 & 113.04k \\
\hline
$10\%$ & 64 & 30 & 30 & 5 & 170 & 107.34k \\
\hline
\end{tabular}
\label{sup1}
% \vspace{-10pt}
\end{table}

\begin{table}[t]\scriptsize
\centering
\caption{Setting of RefCOCO+. Indications follow the Table.\ref{sup1}.}
\setlength{\tabcolsep}{2mm}
\begin{tabular}{c|c|c|c|c|c|c}
\hline 	
$\rm PR$ & $\rm BS$ & $\rm BI~Ep$ & $\rm ACT~Ep$ & $\rm ACT~It$ & $\rm LRD~Ep$ & $\rm Tr~St$  \\
\hline
$0.1\%$ & 8 &  400 & 150 & 11 & 2000 & 105.00k \\
\hline
$1\%$ & 16 & 80 & 30 & 11 & 400 & 104.34k\\
\hline
$5\%$ & 32 & 30 & 30 & 5 & 170 & 112.56k \\
\hline
$10\%$ & 64 & 30 & 30 & 5 & 170 & 107.01k \\
\hline
\end{tabular}
\label{sup2}
% \vspace{-10pt}
\end{table}

\begin{table}[t]\scriptsize
\centering
\caption{Setting of RefCOCOg-umd. Indications follow the Table.\ref{sup1}.}
% \vspace{-10pt}
\setlength{\tabcolsep}{2mm}
\begin{tabular}{c|c|c|c|c|c|c}
\hline 	
$\rm PR$ & $\rm BS$ & $\rm BI~Ep$ & $\rm ACT~Ep$ & $\rm ACT~It$ & $\rm LRD~Ep$ & $\rm Tr~St$  \\
\hline
$0.1\%$ & 8 &  400 & 150 & 11 & 2000 & 70.00k \\
\hline
$1\%$ & 16 & 80 & 30 & 11 & 400 & 69.67k \\
\hline
$5\%$ & 32 & 30 & 30 & 5 & 170 & 75.45 k  \\
\hline
$10\%$ & 64 & 30 & 30 & 5 & 170 & 71.55k \\
\hline
\end{tabular}
\label{sup3}
\end{table}

\section{Additional Implementation Details}
Here, we provide additional details on the implementation of the main content. We adhere to the original model's loss function and hyperparameters. 
For the loss associated with the quantized detection head of TransVG~\cite{deng2021transvg}, we utilize cross-entropy (CE) with a coefficient of 0.1.
In the case of VLTVG~\cite{yang2022improving}, we employ binary cross-entropy with a coefficient of 1 to better match its multi-layer loss. 
Regarding the training hyperparameters, adjustments are made to accommodate variations in training set sizes resulting from different datasets and semi-supervised settings. This ensures that the total number of training steps remains within a reasonable range. Further details and elaboration on these adjustments can be found in Table~\ref{sup1}, Table~\ref{sup2}, and Table~\ref{sup3}.

\section{Limitations}
Due to the current stage of development in neural network interpretability \cite{selvaraju2017grad, Chefer_2021_CVPR, chefer2021generic}, the attribution heatmap has not yet reached perfection and may cause slight discrepancies when calculating faithfulness. In other words, the model's decision-making process may not be accurately represented in the heatmap. This may result in some samples not being selected during sampling, but it does not impact the overall quality of the sampled data. Additionally, the Confidence and Robustness metrics can serve as complementary measures to address shortcomings in the Faithfulness metric.

% \newpage

% {
%     \small
%     \bibliographystyle{ieeenat_fullname}
%     \bibliography{main}
% }

\end{document}